# Mining a Sub-Matrix of Maximal Sum

Vincent Branders, Pierre Schaus, and Pierre Dupont

Université catholique de Louvain - ICTEAM/INGI - Machine Learning Group
Place Sainte Barbe 2, 1348 Louvain-la-Neuve - Belgium
`firstname.lastname@uclouvain.be`

**Abstract.** Biclustering techniques have been widely used to identify homogeneous subgroups within large data matrices, such as subsets of genes similarly expressed across subsets of patients. Mining a *max-sum sub-matrix* is a related but distinct problem for which one looks for a (non-necessarily contiguous) rectangular sub-matrix with a maximal sum of its entries. Le Van et al. [6] already illustrated its applicability to gene expression analysis and addressed it with a constraint programming (CP) approach combined with large neighborhood search (CP-LNS). In this work, we exhibit some key properties of this $\mathcal{NP}$-hard problem and define a bounding function such that larger problems can be solved in reasonable time. Two different algorithms are proposed in order to exploit the highlighted characteristics of the problem: a CP approach with a global constraint (CPGC) and mixed integer linear programming (MILP). Practical experiments conducted both on synthetic and real gene expression data exhibit the characteristics of these approaches and their relative benefits over the original CP-LNS method. Overall, the CPGC approach tends to be the fastest to produce a good solution. Yet, the MILP formulation is arguably the easiest to formulate and can also be competitive.

## 1 Introduction

Gene expression data is typically represented as a large matrix of gene expression levels across various samples. The study of such data is a valuable tool to improve the understanding of the underlying biological processes. For example, biomarker discovery aims at finding indicators of a disease or the physiological state of patients. This problem can be addressed with clustering techniques which perform a grouping of one dimension, either the rows or the columns of the original matrix. Yet it is known that breast cancer, for example, exhibits distinct subtypes [11,12]. In other words, specific genes exhibit activation patterns only in a specific group of patients. Biclustering techniques, or co-clustering, identify specific subsets of rows and of columns which jointly form homogeneous entries [8].

In the present work, we focus on a related but different mining task. One looks in particular for subsets of rows and of columns with globally high values. In the context of gene expression analysis, the objective is to find a subset of genes which are relatively highly expressed among a subset of patients, even though some entries might depart from this pattern. Formally, one looks for a rectangular, and non necessarily contiguous, sub-matrix of a large matrix with a maximal sum of the selected entries.

This *max-sum sub-matrix* problem is closely related to the *maximal ranked tile* mining problem studied by Le Van et al. [6]. In the later case, prior to the search of a sub-matrix, each matrix entry is replaced by its rank across its particular row. In other words, the maximal ranked tile mining is equivalent to the max-sum sub-matrix for which the matrix entries are discrete ranks along the rows. Since the combinatorial optimization algorithms to solve such problems are actually not specific to discrete entries we address here the slightly more general setting of continuous entries.

Our main contributions are 1) the study of the *max-sum sub-matrix* problem while exhibiting some of its key properties and the definition of an upper bound easy to compute in order to speed up the search for a solution; 2) the implementation of two additional algorithms making use of these properties : a CP approach with a global constraint (CPGC) and mixed integer linear programming (MILP); 3) practical experiments conducted both on synthetic and real gene expression data showing that the original CP-LNS method can be largely outperformed; 4) the study of the relative benefits of the proposed methods across various problem instances.

## 2 Problem

### 2.1 Statement

**Definition 1 (The Max-Sum Sub-Matrix Problem).** *Given a matrix $\mathcal{M} \in \mathbb{R}^{m \times n}$ consisting of $m$ rows and $n$ columns, let $\mathcal{R} = \{1, \ldots, m\}$ and $\mathcal{C} = \{1, \ldots, n\}$ be index sets for rows and for columns respectively, find the max-sum sub-matrix $(I^*, J^*)$, with $I^* \subseteq \mathcal{R}$ and $J^* \subseteq \mathcal{C}$, such that:*

$$(I^*, J^*) = \operatorname*{argmax}_{I \subseteq \mathcal{R}, J \subseteq \mathcal{C}} f(I, J) = \operatorname*{argmax}_{I \subseteq \mathcal{R}, J \subseteq \mathcal{C}} \sum_{i \in I, j \in J} \mathcal{M}_{i,j} \ . \tag{1}$$

The *objective function* $f(I, J)$ is the sum of the entries of a sub-matrix $(I, J)$. The maximization term rewards, respectively penalizes, matrix entries with positive, respectively negative, values. The problem is interesting only for matrices with positive and negative entries. Otherwise the optimal solution is trivially identified.

The max-sum sub-matrix problem is $\mathcal{NP}$-*hard* from a simple reduction[1] of the Maximum edge Weight Biclique Problem (MWBP) [3].

---

[1] Essentially by considering the rows and columns of the matrix as the two sets of nodes of a bipartite graph.

### 2.2 Properties

For a defined subset of columns $J$, the objective function can be formulated as $f(I, J) = \sum_{i \in I} r_i$ with $r_i$ being the contribution of the row $i$:

$$r_i = \sum_{j \in J} \mathcal{M}_{i,j} \ . \tag{2}$$

This is important as the search can be limited to one dimension through independent computation of the contributions along the other dimension. Indeed, for any of the two dimensions being fixed, optimization along the other dimension is trivial since it amounts to select only the subset of entries whose contribution is positive. For a fixed subset of columns $J \subseteq \mathcal{C}$, the subset of rows $I_J^* \subseteq \mathcal{R}$ that maximizes the objective value is greedily identified as:

$$I_J^* = \underset{I \subseteq \mathcal{R}}{\operatorname{argmax}} \sum_{i \in I, j \in J} \mathcal{M}_{i,j} = \{i \in \mathcal{R} \mid r_i \geq 0\} \ . \tag{3}$$

In the gene expression analysis context with order(s) of magnitude more rows (the genes) than columns (the samples), one typically search for a subset of columns and greedily select the associated optimal subset of rows.

The search space of the max-sum sub-matrix problem contains $2^{|\mathcal{R}|} \times 2^{|\mathcal{C}|}$ *feasible solutions* that are rectangular sub-matrices $(I, J)$ of the original matrix, $(I \subseteq \mathcal{R}, J \subseteq \mathcal{C})$. Thanks to the independent contribution along one dimension, the number of feasible solutions to be explicitly evaluated is thus reduced to $2^{|\mathcal{C}|} \times |\mathcal{R}|$.

### 2.3 Related Work

*Biclustering* techniques address the problem of finding sub-matrices that satisfy some definition of homogeneity since entries grouped together into biclusters typically have similar values. A comprehensive review can be found in [8]. There is no assumption of homogeneity in the max-sum sub-matrix problem but rather one looks for a rectangular sub-matrix with an overall maximal sum.

*The maximum (contiguous) subarray* problem introduced in [2] identifies a subarray of maximal sum from an array. For a one-dimensional array, this problem can be solved in linear time by Kadane's algorithm [2]. Cubic and sub-cubic time complexity algorithms have been proposed in the two-dimensional case [2,14,15]. This problem is however simpler than the max-sum sub-matrix since the selected sub-matrix is constrained to be formed of contiguous rows and contiguous columns from the original matrix.

*The maximum ranked tile* mining problem has been introduced in [6]. As discussed in Sect. 1, this is a special case of the max-sum sub-matrix problem for which the matrix entries are discrete ranks, corresponding to a permutation of column indices on each row. Both problems are examples of subgroup discovery which aims at finding subsets of a dataset that show an interesting behavior with respect to a certain property of interest [1].

## 3 Optimization Approaches

### 3.1 Boolean Decision Vectors

The max-sum sub-matrix problem can be modeled with two vectors of boolean decision variables: $T = (T_1, \ldots, T_m)$ for the rows and $U = (U_1, \ldots, U_n)$ for the columns with $T_i \in \{0, 1\}$ and $U_j \in \{0, 1\}$. A sub-matrix $(I, J)$ is defined by $I = \{i \in \mathcal{R} \mid T_i = 1\}$ and $J = \{j \in \mathcal{C} \mid U_j = 1\}$. The problem consists in assigning a value to each variable of $T$ and $U$.

Let us denote by $U^1 = \{j \in \mathcal{C} \mid U_j = 1\}$ the selected columns, $U^0 = \{j \in \mathcal{C} \mid U_j = 0\}$ the unselected ones and by $U^? = \{j \in \mathcal{C} \mid U_j = \{0, 1\}\}$ the undecided ones.

### 3.2 Constraint Programming

Constraint Programming (CP) is a flexible programing paradigm that is capable of solving optimization problems. As a declarative approach, it only requires to model the problem and, by using existing solvers, let it search and find solutions. A model is defined as a constraint satisfaction problem $CSP = (V, D, C)$ where $V$ is the set of variables, $D$ is their respective domains, and $C$ is a set of constraints defined over the variables. A feasible solution is an assignment of the variables to values of their domains such that all constraints are satisfied.

Constraints are exploited to iteratively reduce the domains of variables. Such *constraint propagation* reduces the number of variable assignments to consider. Once all unfeasible values are removed from the domains of variables, the *fix-point* of the propagation is reached. Then the solver selects a variable $X \in V$ that is unbound and recursively calls the solver while assigning a value to this variable. Through exploration of a depth-first-search tree (DFS), the solver either reaches a solution or backtrack when the domain of variables becomes empty.

Efficient backtracking is achieved through trailing, which is a state management strategy that facilitates the restoration of the computation state to an earlier version, effectively *undoing* changes that were imposed since then. Knuth [5] attributes to Floyd [4] the first formulation of this mechanism and its usage in a backtracking algorithm. The *trail* exposes two methods: `pushState` and `popState` to respectively time-stamp the current state and restore it. Its implementation is captured in terms of two simple stacks. The first stack holds entries to undo, the second one holds the frame sizes stacked between two consecutive call to `pushState`. Trailing enables the design of reversible objects defined on the trail.

**CP-LNS.** The max-sum sub-matrix problem can be modeled as:

$$\text{maximize} \sum_{i \in \mathcal{R}} T_i \times r_i \ , \forall i \in \mathcal{R} : T_i = 1 \Leftrightarrow r_i \geq 0 \ . \tag{4}$$

The objective is to maximize the sum of rows contributions which are formalized as $r_i = \sum_{j \in \mathcal{C}} U_j \times \mathcal{M}_{i,j}$. Each row with positive (respectively negative) contribution is constrained to be selected (respectively unselected). This provide strong filtering during the search.

A speed up of the search, in the form of a large neighborhood search (LNS), has also been proposed by Le Van et al. [6]. Through modification of the assignment of a large number of variables, LNS avoids being stuck in a region of the search tree by restarting the exploration in other regions. This comes at the cost of losing the possibility to prove optimality. In the present problem, half of the column variables are uniformly selected for the search while leaving the others unchanged.

**CP Global Constraint.** We propose to improve the model by Le Van et al. [6] by designing an efficient algorithm encapsulated inside a global constraint that captures the whole problem. The pseudo-code is given in Algorithm 1. The call to the bounding and filtering procedure has been made explicit. In practice, the lines 11 to 14 would be encapsulated in the global constraint triggered by the fix-point algorithm. The key ingredients of our approach are:

- A feasible solution at each node of the search tree (`evaluate()`).
- An efficient bounding procedure (`upperBound()`).
- An efficient procedure to filter the domains (`filter()`).

*Feasible solution at each node of the search tree.* CP usually updates its feasible solution and best so far lower bound in the leaf-node of the search tree, that is when every variable of the problem is bound. One can observe that for the max-sum sub-matrix problem, any partial assignment of the variables can be extended implicitly as a complete solution for which the unbound variables would be set to 0 (i.e. $U^?$ variables are considered unselected). There is thus no need to wait that every variable is bound to evaluate the solution and possibly update the best so far lower bound. The value of the objective function of the feasible solution is computed in the `evaluate()` method as the sum of the positive rows contributions of the partial solution (see eq. (2) and (3) where $U^1 = J$): $f(I^*_{U^1}, U^1) = \sum_{i \in \mathcal{R}} \max(0, r_i)$.

*An efficient bounding procedure.* CP uses a branch and bound depth-first-search to avoid the exploration of proven sub-optimal solution. The branch and bound performances depend on the strength and efficiency of the procedure to compute the upper bound. We design simple yet efficient bounding procedure for the max-sum sub-matrix problem. Intuitively one computes an upper bound on the row contribution of each row and sum up all the positive bounds on the rows. The upper bound on the contribution of a row is the sum of the matrix entries in the selected columns plus the sum of the positive entries in the unbound columns. One simply computes the upper bound as:

$$g(P) = g(U^1, U^0, U^?) = \sum_{i \in \mathcal{R}} \max(0, r_i + \sum_{j \in U^?} \max(0, \mathcal{M}_{i,j})) \ . \tag{5}$$

The bound is admissible but not tight as it may optimistically evaluate the objective ($g(P) \geq f(P)$). Indeed, it relies on a per-line relaxation of the problem, each selecting a possibly different set of columns.

The `upperBound()` method is an implementation of the proposed upper bound using reversible double to store the incremental modifications of the partial contribution of the rows. Using a reversible sparse-set $\mathcal{T}$ for the row variables allows an efficient exclusion or inclusion of the rows through descent or backtrack [13]. Indeed, as soon as the bound of the row contribution becomes negative it should not be considered in the subsequent nodes of the search tree. The number of rows to consider goes from $\theta(|R|)$ to $O(|R|)$.

*An efficient filtering procedure.* The `filter()` method evaluates the upper bound result for two one-step look-ahead scenarios: if column $j$ would be selected, this look-ahead upper bound is denoted $ub_j^{\in}$, or if $j$ would not be selected, denoted $ub_j^{\notin}$. Then, any column $j$ with $ub_j^{\in} \leq best$ is discarded as inclusion of the column can only lead to worst solution than the incumbent. Similarly, any column $j$ with $ub_j^{\notin} \leq best$ is included. The time complexity for computing all the look-ahead upper bounds is in $O(|\mathcal{T}| \times |U^?|)$ by taking advantage of the fact that the look-ahead bound of each line for each column can be obtained in $O(1)$ from $r_i^{\text{ub}}$.

### 3.3 Mixed Integer Linear Programming

Mixed Integer Linear Programming (MILP) [9] involves the optimization of a linear objective function, subject to linear constraints. Some or all of the variables are required to be integer. A MILP solver explores a branch and bound tree using linear-programming (LP) bounds at each node of the search tree.

As for CP, there are advantages to using MILP. It represent a natural alternative to constraint programming as the problem is modeled with integer decision variables.

**Model.** The max-sum sub-matrix problem can be formulated as a linear model using "big $M$" constants. Equations (7) and (8) linearize $r_i^+ = \max(0, r_i)$ with $r_i$ being the sum of the selected entries of row $i$.

$$\text{maximize} \sum_{i \in \mathcal{R}} r_i^+ \;, \tag{6}$$

$$\text{s.t. } r_i^+ \leq T_i \times M, \quad \forall i \in \mathcal{R} \;, \tag{7}$$

$$r_i^+ \leq r_i + (1 - T_i) \times M, \quad \forall i \in \mathcal{R} \;, \tag{8}$$

$$r_i = \sum_{j \in \mathcal{C}} \mathcal{M}_{i,j} \times U_j, \quad \forall i \in \mathcal{R} \;. \tag{9}$$

In order to avoid rounding errors and ill conditioned matrices, the big $M$ constants can be replaced by $\sum_{j \in \mathcal{C}} \max(0, \mathcal{M}_{i,j})$ and $-\sum_{j \in \mathcal{C}} \min(0, \mathcal{M}_{i,j})$, respectively in equations (7) and (8).

**Algorithm 1:** CP Global Constraint

```
 1  best ← −∞  // best so far objective
 2  trail
 3  r: array[m] rev-double  // rows partial sums
 4  T: rev-set ← {1, . . . , n}  // candidate rows

 5  Method dfs()
 6      if U^? ≠ φ then
 7          j ← select j ∈ U^?
 8          foreach v ∈ {0, 1} do
 9              trail.pushState()
10              D(U_j) ← v
11              best ← max(best, evaluate())
12              ub ← upperBound()
13              if ub > best then
14                  filter()
15                  dfs()
16              trail.popState()

17  Method evaluate(): double
        // evaluate objective
18      for j ∈ U^? do
19          if D(U_j) ≠ {0, 1} then
20              U^? ← U^? \ j
21              if D(U_j) = 1 then
22                  r_i ← r_i + M_{i,j}, ∀i ∈ T
23      return ∑_{i∈T} max(0, r_i)

24  Method upperBound(): double
25      ub ← 0
26      for i ∈ T do
27          r_i^{ub} ← r_i + ∑_{j∈U^?} max(0, M_{i,j})
28          if r_i^{ub} > 0 then ub ← ub + r_i^{ub}
29          else T ← T \ i   // r_i always ≤ 0
30      return ub

31  Method filter(): double
        // remove impossible values
32      ub_j^∈ ← 0 ∀j ∈ U^?
33      ub_j^∉ ← 0 ∀j ∈ U^?
34      for i ∈ T do
35          for j ∈ U^? do
36              if M_{i,j} > 0 then
37                  if r_i^{ub} − M_{i,j} > 0 then
38                      ub_j^∉ ← ub_j^∉ + r_i^{ub} − M_{i,j}
39                  ub_j^∈ ← ub_j^∈ + r_i^{ub}
40              else
41                  ub_j^∉ ← ub_j^∉ + r_i^{ub}
42                  if r_i^{ub} + M_{i,j} > 0 then
43                      ub_j^∈ ← ub_j^∈ + r_i^{ub} + M_{i,j}

44      for j ∈ U^? do
45          if ub_j^∈ ≤ best then D(U_j) ← 0
46          if ub_j^∉ ≤ best then D(U_j) ← 1
```

## 4 Experiments

This section describes experiments conducted to assess the relative performances of three algorithms to solve the max-sum sub-matrix problem. CP-LNS denotes the method proposed by Le Van [6]. The other algorithms are specific methods proposed in the present work: a constraint programming with a global constraint (CPGC) and a mixed integer programming (MILP) solution.

These algorithms are first compared on data matrices which are generated in a controlled setting. Experiments on the breast cancer gene expression data used in [6] are reported next. The main criterion to assess the performance of the various methods is the computing time to solve a particular problem instance. This is technically assessed through an any-time profile defined below.

All algorithms have been implemented in the Scala programming language (2.11.4). Each run is executed with a single thread on a MacBook Pro (OS version 10.10.5) laptop (Intel i7-2720 CPU @ 2.20-3.30GHz, 4GB RAM per run). Constraint programming implementations are based on the latest version of OscaR [10] and MILP is based on the latest version of Gurobi (7.0.2).

The code and datasets are available at `https://bitbucket.org/vbranders/maxsumsubmatriximplementation`.

### 4.1 Synthetic Data

We follow a similar protocol as in [6]. Synthetic data are generated by implanting a sub-matrix $(I, J)$ of interest in a larger matrix $\mathcal{M} = (\mathcal{R}, \mathcal{C})$ made of $m$ rows and $n$ columns. The implanted sub-matrix $(I, J)$ forms a specific selection of rows and columns chosen at random. For each row index (from 1 to $m$) and each column index (from 1 to $n$) of $\mathcal{M}$, a binary variable is sampled from a Bernoulli distribution $B(p)$ and the associated row or column is included in the sub-matrix $(I, J)$ if $B(p) = 1$. Hence, $I = \{i \in \mathcal{R} \mid B(p) = 1\}$ and $J = \{j \in \mathcal{C} \mid B(p) = 1\}$. Next, the full matrix $\mathcal{M}$ is generated according to two normal distributions, $\mathcal{N}(1, 1)$ whenever the particular entry belongs to the implanted sub-matrix, and $\mathcal{N}(-3, 1)$ otherwise.

$$\forall i \in \mathcal{R}, \forall j \in \mathcal{C} : \mathcal{M}_{i,j} = \begin{cases} \sim \mathcal{N}(1, 1) & \text{if } i \in I \text{ and } j \in J \ , \\ \sim \mathcal{N}(-3, 1) & \text{otherwise} \ . \end{cases} \quad (10)$$

Such a generation protocol favors the occurrence of higher values in the implanted sub-matrix and lower values elsewhere. Yet, given the standard deviations chosen equal to 1, both ranges of values may overlap. We note that, as in [6], the implanted sub-matrix is not guaranteed to be an optimal solution to the max-sum sub-matrix problem. This generation protocol looks however realistic to define a rectangular (and not necessarily contiguous) sub-matrix of interest in a larger matrix.

Problem instances are generated for various matrix sizes $(m, n)$ and a varying parameter $p$. As $p$ increases, the size of the implanted sub-matrix is expected to increase as well. In the gene expression analysis context, $m$ can easily be two orders of magnitude larger than $n$ and the sub-matrix of interest is typically small as compared to the full matrix. Such cases are included in the controlled experiments reported below but a larger spectrum of problem instances is also considered.

### 4.2 Gene Expression Data

The proposed case study concerns biomarker discovery for breast cancer subtypes using heterogeneous molecular data types. For a biological analysis and interpretation of the results, the interested reader is redirected to the work of [6]. The pre-processed data provided by [6] consists of a matrix of $m = 2,211$ rows and $n = 94$ columns. In a first step, the matrix entries are transformed to discrete ranks along each row. A given threshold $\theta \times n$ is then subtracted from each matrix entry.

As $\theta$ increases, the proportion of positive entries decreases and, consequently, a smaller sub-matrix of interest is expected to be found. Hence, the control parameter $\theta$ plays a similar role as the parameter $p$ (from Sect. 4.1) but in an opposite way.

### 4.3 Evaluation

One could assess algorithms performances through runtime or number of feasible solutions. While the former may depend on implementation details, the latter strongly depends on the time spent in each node. We preferred the runtime comparisons as it is a more common approach and we made sure to implement the algorithms in the most comparable fashion.

**Any-Time Profile.** In practice, an important criterion for the user is the time required to solve an instance and the ability to find the best solution within a given budget of time. Using *any-time profiles*, one can summarize these characteristics. The idea behind any-time profiles is that an algorithm should produce as high quality solution as possible at any moment of its running time [7].

In the max-sum sub-matrix problem, a high solution quality corresponds to a sub-matrix of large sum. Performances of each algorithm are aggregated on multiple instances. The solution quality for each algorithm is scaled on a $[0, 1]$ interval and the reported result is the average per-instance solution quality as a function of the time. For each instance, runs not completed in a maximum budget of time $t^{\max}$ are interrupted.

**Definition 2 (Max-Sum Sub-Matrix Any-Time Profile).** *Let $f(algo, inst, t)$ be the objective value of the best solution found so far by an algorithm algo for an instance inst at time t. Let $t^{\max}$ be the maximum running time before interrupting an algorithm. The any-time profile of an algorithm is the solution quality $Q_{algo}(t)$ computed on all instances as a function of the time:*

$$Q_{algo}(t) = \frac{1}{|inst|} \sum_{inst} \frac{f(algo, inst, t)}{f(algo^*_{inst}, inst, t^{\max})} \ , \tag{11}$$

*with $algo^*_{inst} = \underset{algo}{\operatorname{argmax}} f(algo, inst, t^{\max})$.*

### 4.4 Results

Figure 1 presents the any-time profile on 50 synthetic data with $10,000$ rows and $p = \{0.05, 0.3, 0.7\}$ for 100 columns (column 1) or $1,000$ columns (column 2) and the any-time profile on breast cancer gene expression data with $2,211$ rows, 94 columns and variable choices of $\theta$ (columns 3 and 4).

**Synthetic Data.** The existing CP-LNS method from [6] is clearly outperformed by other methods. It can barely produce any solution in the allocated time budget. The best approach is CPGC followed by MILP. The reported curves are stopped whenever the proof of optimality is obtained or else the maximal running time is reached. Hence, CPGC also exhibits best results whenever proving optimality is possible in the allocated running time.

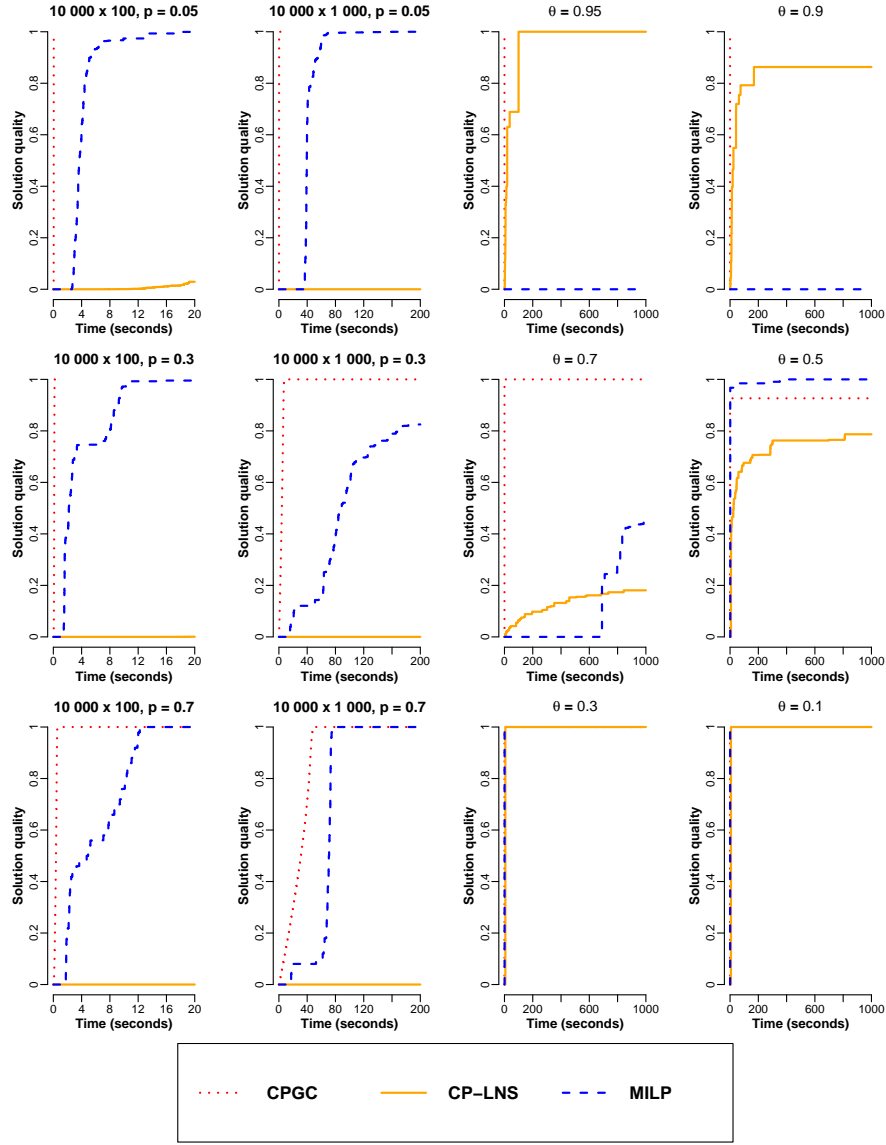

**Fig. 1.** Any-time profiles of constraint programming with a global constraint (CPGC), the method proposed by Le Van et al. [6] (CP-LNS) and mixed integer linear programming (MILP). Columns 1 and 2: Curves corresponds to the average solution quality over all instances as a function of time (in seconds). Results are computed on 50 instances with $10,000$ rows, $100$ (column 1) or $1,000$ (column 2) columns, a variable $p$ and a maximum running time of $20$ (column 1) or $200$ (column 2) seconds. Columns 3 and 4: Curves corresponds to the solution quality over each problem instance obtained for a specific $\theta$ as a function of the time (in seconds). Results computed on breast cancer gene expression data with $2,211$ rows, $94$ columns and various $\theta$ values for a maximum waiting time of $1,000$ seconds.

**Gene Expression Data.** Each curve corresponds here to the performance of an algorithm on a single instance, the one obtained for a specific choice of $\theta$. On the whole spectrum of instances considered, the clear winner is CPGC. The most interesting instances are those for which $\theta \geq 0.9$ since such settings correspond to small sub-matrices which are more likely to illustrate an interesting biological pattern. In such cases, the best approaches are CPGC and CP-LNS.

## 5 Conclusions and Perspectives

We introduce the *max-sum sub-matrix* problem which consists in finding a (non necessarily contiguous) rectangular sub-matrix in a large matrix whose sum is maximal. This problem is originally motivated, in the context of gene expression analysis, by the search of a subset of highly expressed genes in a specific subset, to be found, of relevant samples exhibiting such a pattern. A close variant of this problem, known as *maximal ranked tile* mining problem, has already been studied and tackled with constrained programming (CP) combined with large neighborhood search (CP-LNS) [6].

We present here key properties of the max-sum sub-matrix problem to speed up the search for a solution and we propose two additional algorithms to solve it. Experiments reported both on synthetic data and the original gene expression data used in [6] illustrate the benefits of our proposed methods. In particular, a CP approach with a global constraint (CPGC) is the most effective one in a large spectrum of problem instances. Overall, the CPGC method is also best at proving optimality when such proof can be obtained within the allocated budget of computing time.

The second approach proposed here relies on mixed integer linear programming (MILP). It is arguably the simplest to formulate and to address with a standard solver for such problems. It is competitive with the other methods and largely outperforms CP-LNS as well in our controlled experiments. It exhibits however some performance degradation on some instances from gene expression data, most likely as a consequence of the specific relaxation it is based on.

Related biclustering techniques identify multiple sub-matrices. The max-sum sub-matrix problem could be extended to such a setting by iteratively selecting max-sum sub-matrices penalizing the previous discovered rows and columns to avoid the selection of already identified sub-matrices.

The max-sum sub-matrix mining problem could be extended to a supervised classification setting. For example, in gene expression analysis, one typically wants to find genes (rows) that allows to discriminate between two conditions. In other words, the columns could be *a priori* labeled according to two conditions. The objective can then be to identify a subset of rows that are maximally relevant to discriminate between subsets of samples from different conditions. This could be encoded in a larger matrix for which columns represent pairs of columns in either conditions from the original matrix and the value stored is interpreted as a distance value for a particular gene across both conditions.

The max-sum sub-matrix problem could also be applied to outlier detection and/or biclustering. For example, using an appropriate data transformation, entries that are close to the mean or to the median could be mapped to relatively large positive entries. Similarly, entries far away from the mean would be mapped to low values. Consequently a sub-matrix of maximal sum would correspond to subsets of rows and of columns exhibiting similar entries. Explicit comparisons to existing biclustering algorithms could be considered in such a setting.

# References

1. Atzmueller, M.: Subgroup discovery. Wiley Interdisciplinary Reviews: Data Mining and Knowledge Discovery 5(1), 35–49 (2015)
2. Bentley, J.: Programming pearls: algorithm design techniques. Communications of the ACM 27(9), 865–873 (1984)
3. Dawande, M., Keskinocak, P., Tayur, S.: On the biclique problem in bipartite graphs (1996)
4. Floyd, R.W.: Nondeterministic algorithms. Journal of the ACM (JACM) 14(4), 636–644 (1967)
5. Knuth, D.E.: The art of computer programming: Volume 4B, Combinatorial Algorithms: Part 2, Backtrack Programming, vol. 4B. Adison-Wesley (2016)
6. Le Van, T., Van Leeuwen, M., Nijssen, S., Fierro, A.C., Marchal, K., De Raedt, L.: Ranked tiling. In: Joint European Conference on Machine Learning and Knowledge Discovery in Databases. pp. 98–113. Springer (2014)
7. López-Ibánez, M., Stützle, T.: Automatically improving the anytime behaviour of optimisation algorithms. European Journal of Operational Research 235(3), 569–582 (2014)
8. Madeira, S.C., Oliveira, A.L.: Biclustering algorithms for biological data analysis: a survey. IEEE/ACM Transactions on Computational Biology and Bioinformatics (TCBB) 1(1), 24–45 (2004)
9. Nemhauser, G.L., Wolsey, L.A.: Integer programming and combinatorial optimization. Wiley, Chichester. GL Nemhauser, MWP Savelsbergh, GS Sigismondi (1992). Constraint Classification for Mixed Integer Programming Formulations. COAL Bulletin 20, 8–12 (1988)
10. OscaR Team: OscaR: Scala in OR (2012), available from https://bitbucket.org/oscarlib/oscar
11. Parker, J.S., Mullins, M., Cheang, M.C., Leung, S., Voduc, D., Vickery, T., Davies, S., Fauron, C., He, X., Hu, Z., et al.: Supervised risk predictor of breast cancer based on intrinsic subtypes. Journal of clinical oncology 27(8), 1160–1167 (2009)
12. Perou, C.M., Sørlie, T., Eisen, M.B., van de Rijn, M., Jeffrey, S.S., Rees, C.A., Pollack, J.R., Ross, D.T., Johnsen, H., Akslen, L.A., et al.: Molecular portraits of human breast tumours. Nature 406(6797), 747–752 (2000)
13. de Saint-Marcq, V.l.C., Schaus, P., Solnon, C., Lecoutre, C.: Sparse-sets for domain implementation. In: CP workshop on Techniques foR Implementing Constraint programming Systems (TRICS). pp. 1–10 (2013)
14. Takaoka, T.: Efficient algorithms for the maximum subarray problem by distance matrix multiplication. Electronic Notes in Theoretical Computer Science 61, 191–200 (2002)
15. Tamaki, H., Tokuyama, T.: Algorithms for the maximum subarray problem based on matrix multiplication. In: SODA. vol. 1998, pp. 446–452 (1998)